\documentclass[letterpaper, 10 pt, conference]{ieeeconf}  

\IEEEoverridecommandlockouts                              
\usepackage{cite}
\usepackage{graphics} 
\usepackage{epsfig} 
\usepackage{epstopdf}
\graphicspath {{./img/}}
\usepackage{mathptmx} 
\usepackage{times} 
\usepackage{amsmath} 
\usepackage{amssymb}  
\usepackage{gensymb}
\usepackage{multirow}
\usepackage{picinpar}
\usepackage{multicol}
\usepackage{stfloats}
\usepackage{caption}
\usepackage{graphicx}
\usepackage{subfigure}
\usepackage{wasysym}
\usepackage{textcomp}
\usepackage{float}
\usepackage{bm}

\usepackage[colorlinks,linkcolor=red,anchorcolor=blue,citecolor=green]{hyperref}

\usepackage{color}

\usepackage{array}
\usepackage{mdwmath}
\usepackage{mdwtab}
\usepackage{eqparbox}

\usepackage{xcolor,soul,framed} 
\usepackage{url}
\usepackage{subfigure}
\usepackage{diagbox}
\usepackage{booktabs}
\usepackage{amsfonts}
\usepackage{etoolbox}

\newcommand{\tabincell}[2]{\begin{tabular}{@{}#1@{}}#2\end{tabular}} 

\makeatletter
\patchcmd{\@makecaption}
  {\scshape}
  {}
  {}
  {}
\makeatother
\def\tablename{Table}

\usepackage{algorithm}
\usepackage{algorithmic}
\usepackage{caption}

\newlength\myindent
\begin{document}

\title{FA-Harris: A Fast and Asynchronous Corner Detector for Event Cameras}

\author{Ruoxiang~Li$^{1}$, Dianxi~Shi$^{2,3,*}$, Yongjun Zhang$^{2}$, Kaiyue~Li$^{1}$, Ruihao~Li$^{2,3,*}$
  \thanks{$^{1}$ College of Computer, National University of Defense Technology, Changsha 410073, China}
  \thanks{$^{2}$ Artificial Intelligence Research Center (AIRC),
National Innovation Institute of Defense Technology (NIIDT), Beijing 100166, China}
  \thanks{$^{3}$ Tianjin Artificial Intelligence Innovation Center (TAIIC), Tianjin 300457, China}
  \thanks{$^{*}$ Corresponding author: \href{mailto:dxshi@nudt.edu.cn}{dxshi@nudt.edu.cn} \href{mailto:liruihao2008@gmail.com}{liruihao2008@gmail.com}}
  }

  


\maketitle

\begin{abstract}

Recently, the emerging bio-inspired event cameras have demonstrated potentials for a wide range of robotic applications in dynamic environments. In this paper, we propose a novel fast and asynchronous event-based corner detection method which is called FA-Harris. FA-Harris consists of several components, including an event filter, a Global Surface of Active Events (G-SAE) maintaining unit, a corner candidate selecting unit, and a corner candidate refining unit. The proposed G-SAE maintenance algorithm and corner candidate selection algorithm greatly enhance the real-time performance for corner detection, while the corner candidate refinement algorithm maintains the accuracy of performance by using an improved event-based Harris detector. Additionally, FA-Harris does not require artificially synthesized event-frames and can operate on asynchronous events directly. We implement the proposed method in C++ and evaluate it on public Event Camera Datasets. The results show that our method achieves approximately 8$\times$ speed-up when compared with previously reported event-based Harris detector, and with no compromise on the accuracy of performance.

\end{abstract}


\section{Introduction}
\label{sec:introduction}
Simultaneous Localization and Mapping (SLAM) is widely used in mobile robots and AR/VR scenery. It can construct a map of unknown environment and simultaneously perform 6-DoF tracking in the map. In the past decade, visual SLAM has achieved remarkable success with many prominent implementations of SLAM systems\cite{mur2015orb, engel2018direct}. However, the existing visual SLAM algorithms still have many limitations in challenging environments. 
When 
faced with high-speed movements (easily causing motion blur) or high dynamic range scenes, the existing systems can hardly demonstrate robust performance\cite{cadena2016past}.


With the development of neuromorphic imaging technology and silicon retinas technology, the new emerging dynamic vision sensor (namely, event camera) has shed light on potentials for robotics and computer vision applications. Intrinsically different from traditional cameras, the event camera (e.g. \cite{brandli2014240}) responds to pixel-wise brightness changes at a high time resolution (microseconds) and transmits asynchronously events with very low latency (microseconds), and it has a high dynamic range and is robust to motion blur.

\begin{figure} [t]
  \centering 
   \subfigure{ 
    \includegraphics[width=3.42in]{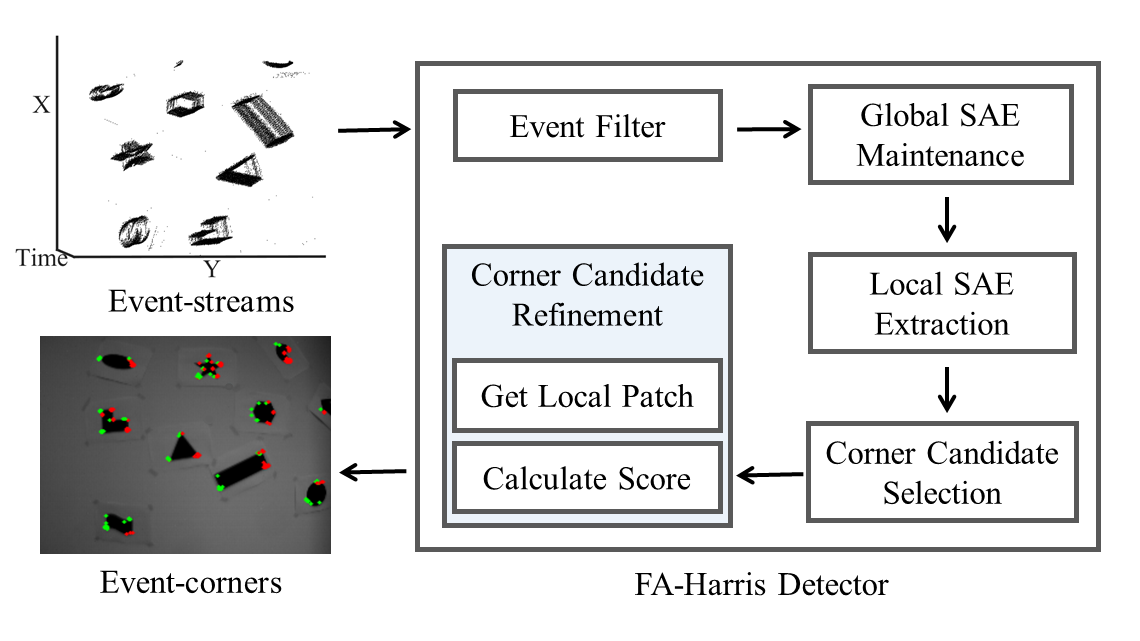} 
  }
  \captionsetup{font={small}}
  \caption{System overview of the proposed FA-Harris detector. It consists of an event filter, a global SAE maintaining unit, a corner candidate selecting unit and a corner candidate refining unit. The intensity-image is used for visualization of the corner detection results in this figure.}
  \label{img:FA-Harris}
\end{figure}

Based on the demonstrated astonishing performance in dynamic environments, the event camera provides the possibility to overcome the bottleneck in visual SLAM. However, the previous algorithms can't be used for event cameras directly. How to deal with asynchronous event-streams is still an open question. New algorithms have to be designed to deal with asynchronous event-streams rather than intensity-frame sequences. In particular, feature extraction is a basic and significant component for feature-based SLAM methods\cite{mur2015orb}. Feature association and visual odometry are both based on it. Driven by the demand for an efficient feature detection method, we present a novel fast and asynchronous corner detector (called FA-Harris) for event cameras in this paper. As shown in Fig. \ref{img:FA-Harris}, our FA-Harris detector consists of an event filter, a global SAE \cite{benosman2014event} maintaining unit, a corner candidate selecting unit and a corner candidate refining unit. It works directly on asynchronous event-streams instead of artificially synthesized event-frames.



The main contributions of this paper can be summarized as follows:
\begin{itemize}

\item We propose a novel fast and asynchronous event-based corner detection method. It works directly on asynchronous events instead of artificially synthesized event-frames. The G-SAE maintenance and corner candidate selection guarantee the speed-up and asynchrony of feature extraction.

\item We address an efficient corner candidate refinement method. It is inspired from Harris detector and operates on G-SAE. The corner candidate refinement computes the candidate score and determines whether the input candidate is a corner.

\item We implement our algorithm in C++ and evaluate it on public Event Camera Datasets. The results show that our proposed method achieves approximately 8$\times$ speed-up when compared with previously reported event-based Harris detector, and with no compromise on the accuracy of performance.

\end{itemize}

In the following section, we address the related works about event-based feature extraction. Section III presents the details and implementation of the proposed FA-Harris detector. Section IV gives the method for the collection of ground truth event-corners. Section V presents the evaluational methods and gives the experimental results. Finally, conclusions are drawn in section VI. 

\section{RELATED WORK}
\label{sec:relatedwork}
In computer vision, the corner is one of the most important image features. Corner detection is frequently used in object recognition, detection, and tracking. In recent years, with the appearance of event cameras, several event-based corner detection methods have been presented. The detection methods can mainly be divided into the non-asynchronous methods and asynchronous methods.

\subsection{Non-synchronous Event-based Corner Detection} 

Non-asynchronous event-based corner detection methods mainly rely on artificially synthesized event-frames. The frames are synthesized by asynchronous events with a fixed number \cite{rebecq2017real, vidal2018ultimate} or in a fixed temporal window \cite{zhu2017event, zhu2017eventodometry}. These methods omit the innate asynchronous nature of event cameras, and are not asynchronous. 

In \cite{zhu2017event, zhu2017eventodometry}, they collected the events in a temporal window to integrate event-frames and detected Harris corners on event-frames. The size of the temporal window is set based on the lifetime \cite{mueggler2015lifetime}. They then tracked the corners using Expectation-Maximization scheme. Rebecq et al. \cite{rebecq2017real} proposed to synthesize event-frames in spatio-temporal windows with events motion compensation. Afterwards, they detected corners on the compensated event-frames using FAST corner detector \cite{rosten2006machine}, and tracked the corners using pyramidal Lukas-Kanade tracking \cite{baker2004lucas}. A more recent work \cite{vidal2018ultimate} detected FAST corners on both motion-compensated event-frames \cite{rebecq2017real} and intensity-frames. Afterwards, they tracked these feature corners based on both frames using Kanade-Lucas-Tomasi (KLT) tracker \cite{Lucas-1981-15102}. In \cite{tedaldi2016feature, kueng2016low, gehrig2018asynchronous}, they first extracted intensity-corners based on intensity-frames using the Harris detector \cite{harris1988combined}, and then tracked these intensity-corners based on event-streams.

For these methods, due to the use of artificially synthesized event-frames or intensity-frames, motion blur may still occur in high-speed sceneries.

\subsection{Asynchronous Event-based Corner Detection} 
\label{subsec:relatedAsynchronous}

Asynchronous event-based corner detection methods \cite{clady2015asynchronous, vasco2016fast, mueggler2017fast, alzugaray2018asynchronous, alzugaray2018ace} detect event-corners directly on asynchronous event-streams instead of artificially synthesized event-frames or intensity-frames. 


The first asynchronous event-based corner detection method was proposed by Clady et al. \cite{clady2015asynchronous}. This method relies on the use of spatio-temporal properties of moving edges. They first collected events within a spatio-temporal window (X-Y-T coordinate). Then fitting planes could be determined by a local linear regularization through fitting the collected events. An event will be labeled as an event-corner if it belongs to the intersection of two or more fitting planes. 

Vasco et al. \cite{vasco2016fast} presented an adaptation of the original intensity-based Harris corner detector \cite{harris1988combined}. They detected the corners directly on the artificially binary frames. 0 and 1 in binary frames indicate the absence and presence of the event at the corresponding pixel position. The binary frame is updated by the most recent 1000 or 2000 events for the whole image plane.

Inspired by frame-based corner detection techniques, Mueggler et al. \cite{mueggler2017fast} proposed a FAST-like event-based corner detection method which is called eFAST. They detected the event-corners on SAE and reduced event-streams to event-corner streams. Surface of Active Events (SAE) is a map with the timestamp of the most recent event at the corresponding pixel position. Compared to \cite{vasco2016fast}, this asynchronous method detected corners more efficiently. Besides, Mueggler et al. \cite{mueggler2017fast} also improved the event-based Harris corner detection method \cite{vasco2016fast} by choosing the locally newest events rather than in the whole image plane. The improved method is called eHarris.

Alzugaray et al. \cite{alzugaray2018asynchronous} presented a more efficient asynchronous corner detection algorithm which is called Arc*. Arc* is inspired from eFAST and can detect corners more effective. It runs 50$\times$ faster than the eHarris method and 4.5$\times$ faster than the eFAST method. They applied an event filter (referred as eFilter in this paper) to reduce the redundant events in Arc*. 
The filtered event is updated only if the previous event $e_l$ at the same location was triggered beyond a fixed temporal window, or the polarity of $e_l$ differs from the polarity of the incoming one.
The authors also employed the eFilter to eHarris and eFAST, respectively. Compared with the original eHarris and eFAST, the improved eHarris* (eFilter+eHarris) and eFAST* (eFilter+eFast) perform better in computational performance. In \cite{alzugaray2018asynchronous, alzugaray2018ace}, they also achieved tracking directly on the event-streams using a completely asynchronous event-based corner tracker. However, Arc* is a FAST-based method essentially, its performance on accuracy is not as good as Harris-based methods.

\subsection{High-level Tasks based on Event Cameras}

Many high-level tasks based on event cameras have already been developed. Reinbacher et al. \cite{reinbacher_iccp2017} proposed a simultaneous 3-DoF tracking and mapping system for event cameras in real-time. Rebecq et al. \cite{rebecq2018emvs} introduced an Event-based Multi-View Stereo (EMVS) system for a single event camera with a known trajectory. By proposing an event-based space sweep method, the EMVS can construct an accurate semi-dense map. Based on EMVS, an Event-based Visual Odometry (EVO) system was proposed in \cite{rebecq2017evo}. EVO can perform simultaneous 6-DoF tracking and mapping in real-time, and the tracking process follows an image-to-model alignment strategy using the geometric error. More recently, Zhou et al. \cite{zhou2018semi} introduced a 3D reconstruction method using a stereo device integrated with two temporally-synchronized event cameras. The method consists of an energy minimization unit and a probabilistic depth-fusion unit to achieve semi-dense 3D reconstruction. Scheerlinck et al. \cite{Scheerlinck19ral} proposed an asynchronous Harris corner detector by introducing an internal state to encode the convolved image information for each event.


 


In conclusion, the non-asynchronous event-based corner detection method needs to use intensity-frames or artificially synthesize event-frames and can hardly achieve asynchronous detection. For the asynchronous event-based method, eHarris is time-consuming which limits its applications in some tasks. The recently proposed Arc* runs faster and detects more event-corners, however, Arc* mistakenly detects more false event-corners as well, and its performance on accuracy is not as good as Harris-based methods.


\section{METHOD}
\label{sec:method}
Driven by the demand for an efficient event-based corner detector, we present a novel fast and asynchronous event-based corner detection pipeline, namely FA-Harris. The FA-Harris detector firstly reduces the redundant events based on the eFilter \cite{alzugaray2018asynchronous} as described in \ref{subsec:relatedAsynchronous}. The filtered events are then used for global Surface of Active Events (G-SAE) construction and update. Afterwards, event-corner candidates are selected efficiently based on G-SAE to reduce the computational pressure. Finally, the selected event-corner candidates are refined by computing the candidate score based on an improved Harris method. The following section will give the details of the proposed method.



\subsection{G-SAE Construction and Update}
\label{subsec:sae}
Different from the traditional cameras which output intensity-frames at a fixed time interval, event cameras output asynchronous event-streams. An event ($x, y, t, p$) contains the position ($x, y$) in image coordinate, the timestamp $t$ and the polarity $p$ ($\pm 1$). The polarity $p$ indicates the brightness change. To process the event-streams, a widely used method in event-based research is SAE. The SAE is similar to an elevation map, it maps the position of the latest event to its timestamp $SAE:(x,y)_e\in\mathbb{R}{^2} \mapsto t_e\in\mathbb{R}$ in the spatio-temporal coordinate. 



eHarris \cite{vasco2016fast} constructs and maintains a local SAE with size $9\times9$ for each pixel in the image plane with size $M\times N$. In other words, there are $M\times N$ local SAEs needing to be maintained. Once a new event comes, all pixels (totally 81 pixels) in the local SAE will update its corresponding local SAE. This kind of SAE maintenance strategy has to maintain a complex data structure which is time-consuming. In order to improve the real-time performance, we proposed to construct and maintain a global SAE with size $M\times N$. G-SAE saves the newest timestamp of the event at the corresponding pixel. When a new event comes, G-SAE will be updated at the corresponding position of the event. Then the local SAE centered on the event will be extracted from G-SAE with size 9$\times$9. The local SAE is then used for corner candidate selection and refinement. The events of two polarities are handled on two different G-SAEs respectively. The proposed G-SAE construction and update method greatly accelerates the speed of our algorithm. We have evaluated our G-SAE maintenance method and proved its efficency (see \ref{subsec:update} for more details).

\subsection{Corner Candidate Selection}
\label{subsec:selection}

In order to further enhance the real-time performance, we take advantage of a corner candidate selection method \cite{alzugaray2018asynchronous} to subsample the original event-streams. In this way, only the selected corner candidates are used for time-consuming computation and refinement. 

For each new incoming event, G-SAE is updated first, and a local SAE is extracted from G-SAE as introduced in section III-A. The local SAE is centered on the event with size 9$\times$9. For the extracted local SAE, we consider two centered concentric circles with radius 3 and 4, respectively. For both circles, a continuous arc $A$ and its complementary continuous arc $A'$ is a segment with continuous pixels on the circle. The arc starts from the pixel with the highest value (newest timestamp) on the circle. For the inner circle, we try to find an arc $A$ that all elements on it are higher than all other elements on the circle, besides, the range of the arc should lie in $[min,max]=[3,6]$. We can also try to find a complementary $A'$ that the elements on it are lower than all other elements on the circle, and the range of the arc should also lie in $[min,max]=[3,6]$. For the outer circle, the range is set to $[min,max]=[4, 8]$. If such an arc $A$ is found on both circles, the incoming event (centered event) is classified as a corner candidate. 

By this post-processing for the orginal events, many invalid events are removed and the corner candidates are selected. This is significant for FA-Harris to maintain low-latency and asynchrony.


\subsection{Corner Candidate Refinement}
\label{subsec:refinement}
The event-based Harris corner detector is one of the most useful detectors for complex texture. In order to achieve better accuracy performance, we post-process the event-corner candidates by adopting a corner candidate score computation method. The method is based on the original Harris detector. For each corner candidate, we apply the same local SAE with size $9\times9$ as adopted in \ref{subsec:selection}, and use it to generate a local patch. The local patch is an essentially binary matrix contains only 0 and 1. We find the most recent $N$ events ($N=25$) on the local SAE, and label them as 1 in the local patch.

Then the local patch is used to calculate corner candidate score as follows:


\begin{equation}\label{harrismatrix}
  M=\left[ \begin{matrix}A&C\\C&B\end{matrix} \right],
\end{equation}
\begin{equation}\label{harrisscore}
\begin{aligned}
  Score&=det(M)-\alpha\cdot trace^2(M)\\
  &=\lambda_1\lambda_2-\alpha\cdot(\lambda_1+\lambda_2)^2\\
  &=(AB-C^2)-\alpha\cdot(A+B)^2\\
  \end{aligned}
\end{equation}
where $\alpha=0.04$ is a constant, $\lambda_1$,$\lambda_2$ are the eigenvalues of $M$, and 
\begin{equation}\label{abc}
\begin{aligned}
A&=\sum_{x,y}{w(x,y)\cdot I^2_x},\\ B&=\sum_{x,y}{w(x,y)\cdot I^2_y},\\ C&=\sum_{x,y}{w(x,y)\cdot I_xI_y}\\
\end{aligned}
\end{equation}
where $I_x$,$I_y$ are the gradients of the patch in x and y direction respectively and they are calculated with a $5\times5$ Sobel kernel, $w(x,y)$ is a Gaussian weighting function with spread $\sigma=1$ pixel as shown below:

\begin{equation}\label{gaussian}
  w(x,y)=\frac{1}{2\pi \sigma^2}e^{\frac{-(x^2+y^2)}{2\sigma^2}}.
\end{equation}

By comparing the computed score with a threshold, the event-corner candidate will be determined whether it is an event-corner.

\subsection{The FA-Harris Algorithm}
\label{subsec:algorithm}
We give the pipeline of our proposed FA-Harris in Algorithm \ref{alg:faharris} as shown below. For a new input event, FA-Harris will first filter it based on eFilter. Then the event will be added to the G-SAE and the G-SAE will be updated. Afterward, FA-Harris will extract a local SAE centered on the event from G-SAE and use it for corner candidate selection. Finally, the candidate score will be computed to determine whether the input event is a corner.

\begin{algorithm}[h]
	\caption{the FA-Harris Algorithm}
	\label{alg:faharris}
	\begin{algorithmic}[1]
		\REQUIRE Event $e=(x,y,t,p)$, G-SAE
		\ENSURE $True$ or $False$
		\STATE eFilter
		\STATE Add $e$ to G-SAE (updating G-SAE)
		\STATE Local SAE extraction centered on $e$
		\IF{$e$ is candidate corner}
		    \STATE Calculate the corner candidate score $s$
			\IF{$s >$ threshold}
				\STATE return $True$
			\ENDIF
		\ENDIF
		\STATE return $False$
	\end{algorithmic}
\end{algorithm}

\section{GROUND TRUTH COLLECTION}
\label{sec:groundtruth}
\begin{figure} [t]
  \centering
   \subfigure{ 
    \includegraphics[width=3.3in]{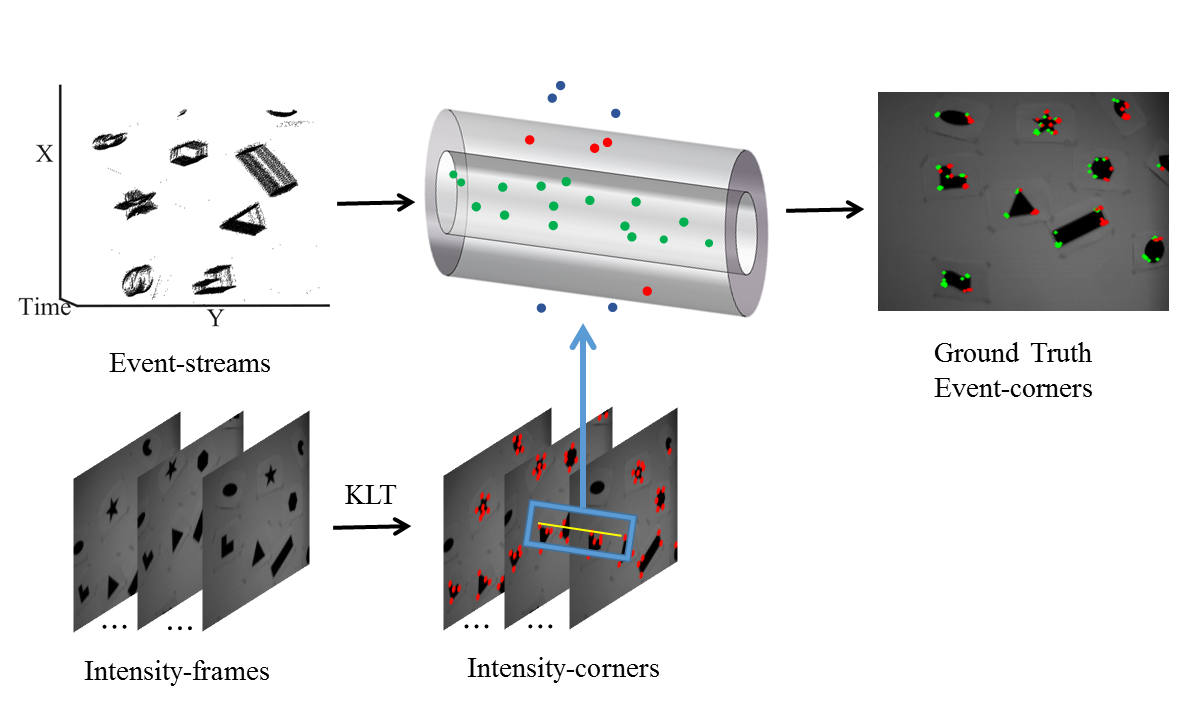} 
  } 
  \captionsetup{font={small}}
  \caption{The Ground Truth Event-Corners Collection Method. The ground truth intensity-corners collection is based on the original Harris detector and KLT tracker. The red intensity-corners in the synchronous intensity-frames are detected and tracked. The oblique cylinder around the optical flow is enlarged (upper middle). The events within 3.5 pixels from the tracked intensity-corners are marked as green and those between 3.5 to 5 pixels are marked as red. We take the green point as true ground truth event-corner, and take the red point as false event-corner. The ground truth event-corners is synthesized on the intensity-image for visualization (upper right). Color means the polarity of event-corner. Green means positive polarity and orange means negative polarity.}
  \label{img:groundtruth}
\end{figure}

\begin{algorithm}[t]
\caption{the Intensity-Corners Collection Algorithm}
\label{alg:groundtruth}
\begin{algorithmic}[1]
	\REQUIRE Intensity-images
	\ENSURE Intensity-corners $C$
	\STATE Read image
	\STATE Detect Harris corners $C_1$ in the image
	\STATE $C = C \cup C_1$
	\WHILE{not the last image}
		\STATE Read image
		\STATE Track corners $C_1$ using KLT tracker resulting $C_2$
		\STATE Detect Harris corners $C_3$ on the image
		\STATE Clear $C_1$
		\STATE Merge $C_3$ and $C_2$ into $C_1$
		\STATE $C = C \cup C_1$
		\STATE Clear $C_2$ and $C_3$
	\ENDWHILE
\STATE return $C$
\end{algorithmic}
\end{algorithm}

Similar with the method presented in \cite{alzugaray2018asynchronous}, we combine the harris detector with KLT tracker, and apply them to synchronous intensity-images to track intensity-corners.


The intensity-corners extracting algorithm is shown in Algorithm \ref{alg:groundtruth}. The algorithm detects intensity-corners $C_1$ in the previous image, and then tracks them using KLT tracker in the current image. We refer to the tracked intensity-corners in current image as $C_2$ which will be further refined in the following process. For the reason that new corners may appear in the current image, we use the Harris detector again in the image and obtain the intensity-corner set $C_3$. $C_3$ and $C_2$ are then merged into one set. If one corner in $C_3$ is less than 5 pixels away from a corner in $C_2$, they are considered as the same corner, and the merged set will be tracked using KLT tracker in the next image. 

In this way, we can estimate the optical flow $\mathbf{v}(v_x, v_y)$ for the intensity-corners. As to the corresponding event-corners between two adjacent images, we use the estimated flow $\mathbf{v}$ and the time interval $\Delta t$ to estimate its position. As shown in Fig. \ref{img:groundtruth}, the same corner on the image plane are tracked and connected with a yellow line segment. Then we take the line segment as the center to construct two oblique cylinders. The radius of the oblique cylinders are 3.5 pixels and 5 pixels, respectively.


For the event-corners fallen inside the cylinder with 3.5 pixels \cite{alzugaray2018asynchronous}, we take them as true event-corners and label them as the ground truth. For the detected event-corners which fallen between 3.5 to 5 pixels, we take them as wrongly detected corners and label them as the false event-corners. Based on the above method, we obtain the ground truth event-corners and use them for the following evaluation.

\section{EXPERIMENTAL EVALUATION}
\label{sec:experiments}
This section introduces our evaluation of the proposed FA-Harris algorithm\footnote{Video: \url{https://youtu.be/v5CcBVkmI6w}} on Event Camera Datasets \cite{mueggler2017event}. The dataset was collected using DAVIS240C \cite{brandli2014240} and contains the asynchronous events, intensity-images, and inertial measurements. We choose the same scenes (\texttt{shapes}, \texttt{dynamic}, \texttt{poster} and \texttt{boxes}) used in \cite{alzugaray2018asynchronous} to achieve qualitative and quantitative evaluation. The collection of the ground truth is introduced in section \ref{sec:groundtruth}. We perform our evaluation work on a laptop equipped with an Intel i7-7700HQ CPU with 2.80GHz and RAM with 16GB.



In the evaluation, we first evaluate the performance of the presented G-SAE. Then the qualitative analysis of our FA-Harris is performed. Following that, we give the accuracy and computational performance of our FA-Harris. We implement the eHarris* and eFAST* method by combining eHarris and eFAST \footnote{\url{https://github.com/uzh-rpg/rpg\_corner\_events}} \cite{mueggler2017fast} with the eFilter \cite{alzugaray2018asynchronous}. The implementations of Arc*\footnote{\url{https://github.com/ialzugaray/arc\_star\_ros}} is provided by \cite{alzugaray2018asynchronous}. All methods are implemented with C++.





\subsection{SAE Updating Performance}
\label{subsec:update}
\begin{table}[t]
	\centering
	\captionsetup{font={small}}
	\caption{Real-time performance of the eHarris* and the G-eHarris* method (defined in \ref{subsec:update}). They use different SAE updating methods, and our G-eHarris* uses G-ASE. The better results are made in bold.}
	\label{table:updatingperformance}
\begin{tabular}{c|c|cc}
		\toprule[1pt]
		\textbf{Scenes} & \textbf{Algorithm} & \tabincell{c}{{\textbf{Updating time}}\\ {\textbf{[s]}}} & \tabincell{c}{{\textbf{Total time}}\\ {\textbf{[s]}}}\\
		\hline
		\multirow{2}*{shapes}   & eHarris*\cite{vasco2016fast}   & 50.64  & 71.85 \\
		\cline{2-4}
		~                       & \textbf{G-eHarris*} & \textbf{0.26}   & \textbf{47.27}\\
		\hline
		\multirow{2}*{dynamic} & eHarris*\cite{vasco2016fast}   & 219.63 & 288.15 \\
		\cline{2-4}
		~                      & \textbf{G-eHarris*} & \textbf{0.79}   & \textbf{147.89}\\
		\hline
		\multirow{2}*{poster}  & eHarris*\cite{vasco2016fast}   & 556.39 & 746.73 \\
		\cline{2-4}
		~                      & \textbf{G-eHarris*} & \textbf{1.94}   & \textbf{375.02} \\
		\hline
		\multirow{2}*{boxes}  & eHarris*\cite{vasco2016fast}   & 640.33 & 812.13 \\
		\cline{2-4}
		~                     & \textbf{G-eHarris*} & \textbf{2.23}   & \textbf{418.82} \\
 		
		\bottomrule[1pt]
	\end{tabular}
\end{table}

In order to evaluate the performance of the G-SAE, we apply the score computation method directly on G-SAE without candidates selection, and use it to detect event-corners with eFilter. Here we call it G-eHarris*. We implement G-eHarris* and evaluate it on Event Camera datasets. 

For a new event, G-eHarris* will update the G-SAE while eHarris* updates the local SAEs as described in \ref{subsec:sae}. To verify the superiority of the G-SAE maintenance method, we test the SAE update performance for both eHarris* and G-eHarris* methods. The results are summarized in Table \ref{table:updatingperformance}. It consists of the SAE updating time and the total time to process the events for all scenes. The results indicates that G-eHarris* is $2\times$ faster than eHarris*, and the ratio of the SAE updating time to the total process time is less than 0.6\% for G-eHarris* compared to about 70\% for eHarris*. 


The results show that our G-SAE maintenance method is significant for the real-time performance of our proposed FA-Harris detector.
The comparison of the accuracy performance about G-eHarris* and eHarris* will be further reported in the following evaluations.


\subsection{Qualitative Analysis}
\label{subsec:analysis}
\begin{figure} [t]
  \centering 
  \subfigure[shapes]{ 
    \includegraphics[width=1.5in]{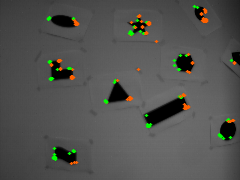} 
    \label{img:shapes}
  }
  \subfigure[dynamic]{ 
    \includegraphics[width=1.5in]{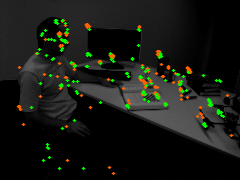} 
    \label{img:dynamic}
  }
  \subfigure[poster]{ 
    \includegraphics[width=1.5in]{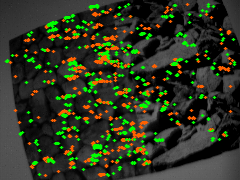} 
    \label{img:poster}
  }
  \subfigure[boxes]{ 
    \includegraphics[width=1.5in]{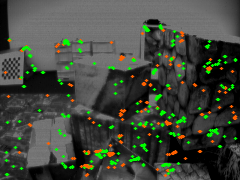} 
    \label{img:boxes}
  }
  \captionsetup{font={small}}
  \caption{The event-corners detected by the FA-Harris detector within 100ms. The intensity-mages are used for visualization. The polarity of event-corners is distinguished by the color. Green means positive polarity and orange means negative polarity. 
}
  \label{img:scene}
\end{figure}

The event-corners detected by the FA-Harris detector for different scenes are shown in Fig. \ref{img:scene}. We synthesize events within 100ms on the intensity-images for visualization. And the colors represent the different polarities of event-corners, green and orange mean positive and negative polarity respectively. Fig. \ref{img:shapes} indicates that our proposed FA-Harris method can detect corners with different angles. In Table \ref{table:reduction}, we report the performance of our proposed method on Reduction Rate (Red.) in percentage. As reported in  \cite{mueggler2017fast}, the reduction rate indicates the ability of a detector that reduce the event-streams to feature-streams. The result shows that FA-Harris reduces the most events than other methods. The number of event-corners Arc* detected is the largest.

\begin{table*}[t]
	\centering
	\captionsetup{font={small}}
	\caption{Performance of different event-corners detection methods on Reduction Rate (Red.).}
	\label{table:reduction}
	\begin{tabular}{c|l|ccccc}
		\toprule[1pt]
		\textbf{Texture} & \textbf{Scenes} & \tabincell{c}{\textbf{eHarris*\cite{vasco2016fast}} \\ \textbf{Red. [\%]}}& \tabincell{c}{\textbf{G-eHarris*} \\ \textbf{Red. [\%]}} & \tabincell{c}{\textbf{eFAST*\cite{mueggler2017fast}} \\ \textbf{Red. [\%]}} & \tabincell{c}{\textbf{Arc*\cite{alzugaray2018asynchronous}} \\ \textbf{Red. [\%]}} & \tabincell{c}{\textbf{FA-Harris} \\ \textbf{Red. [\%]}}\\ 
		\hline
		\multirow{3}*{low} & shapes\_6dof        & 94.84 & 94.96 & 92.74 & 88.34 & \textbf{95.99} \\
		~                  & shapes\_rotation    & 94.90 & 95.02 & 92.14 & 87.69 & \textbf{95.81} \\
		~                  & shapes\_translation & 94.96 & 95.05 & 92.46 & 88.14 & \textbf{95.88} \\
		\hline
		\multirow{3}*{medium} & dynamic\_6dof        & 96.80 & 97.11 & 97.43 & 92.22 & \textbf{98.67} \\
		~                     & dynamic\_rotation    & 96.24 & 96.55 & 97.18 & 91.74 & \textbf{98.46} \\
		~                     & dynamic\_translation & 97.20 & 97.50 & 97.92 & 92.71 & \textbf{98.91} \\
		\hline
		\multirow{3}*{high} & poster\_6dof        & 94.37 & 95.05 & 96.63 & 90.21 & \textbf{98.12} \\
		~                   & poster\_rotation    & 94.18 & 95.01 & 96.47 & 89.92 & \textbf{98.08} \\
		~                   & poster\_translation & 94.85 & 95.58 & 97.09 & 90.89 & \textbf{98.33} \\
		\hline
		\multirow{3}*{high} & boxes\_6dof        & 94.42 & 95.38 & 97.40 & 92.19 & \textbf{98.60} \\		
		~                   & boxes\_rotation    & 94.42 & 94.84 & 97.10 & 91.89 & \textbf{98.40} \\
		~                   & boxes\_translation & 94.32 & 95.39 & 97.41 & 92.25 & \textbf{98.57} \\
		\bottomrule[1pt]
	\end{tabular}
\end{table*}

\subsection{Accuracy Performance}
\label{subsec:detector}

\begin{table}[t]
	\centering
	\captionsetup{font={small}}
	\caption{The \iffalse detailed \fi False Positive Rate (\%) of different event-corners detection methods on different scenes. The best results are made in bold. The 'Overall' result here is the weighted average based on the number of events in each scene.}
	\label{table:fpr}
	\begin{tabular}{l|ccccc}
		\toprule[1pt]
		\diagbox{\textbf{Alg.}}{\textbf{Scene}} & \textbf{shapes} & \textbf{dynamic} & \textbf{poster} & \textbf{boxes} & \textbf{Overall} \\
		\hline
		Arc*\cite{alzugaray2018asynchronous}	   & 29.88 & 18.10 & 13.34 & 11.13 & 14.64 \\
		eFAST*\cite{mueggler2017fast}     & 18.62 & 4.59  & 3.04  & 2.67  & 4.01 \\
 		\hline
		eHarris*\cite{vasco2016fast}   & 13.63 & 8.15  & 7.22  & 7.42  & 7.75 \\
		G-eHarris*  & 13.46 & 7.21  & 6.04  & 6.04  & 6.52 \\
		\textbf{FA-Harris}  & \textbf{11.68} & \textbf{3.28} & \textbf{1.90} & \textbf{1.61} & \textbf{2.58} \\
		\bottomrule[1pt]
	\end{tabular}
\end{table}

Similarly to \cite{alzugaray2018asynchronous}, we calculate the False Positive Rate (\texttt{FPR}) for different methods. The result for \texttt{FPR} is summarized 
in Table \ref{table:fpr}. We consider the events in the oblique cylinder for the evaluation, and only the first 10 seconds of the sequence for all scenes are used in this section\cite{alzugaray2018asynchronous}. For \texttt{FPR}, an event-corner is labelled as true positive (\texttt{TP}) if it is within 3.5 pixels from the tracked intensity-corners or false positive (\texttt{FP}) if it is in the range of 3.5 to 5 pixels. And the event which is not event-corner is labelled as false negative (\texttt{FN}) within 3.5 pixels. The event in the range of 3.5 to 5 pixels is labelled true negative (\texttt{TN}). Then \texttt{FPR} is calculated by \texttt{FPR=FP/(FP+TN)}. The accuracy is calculated by \texttt{TP/(TP+FP)}, and it indicates the proportion of the correct event-corners in the detected event-corners. The results for accuracy evaluation is summarized in Table \ref{table:correct}. As shown in Table \ref{table:correct}, our FA-Harris can detect more correct event-corners and demonstrate better performance in terms of accuracy. 

The False Positive Rate of different event-corners detection methods is shown in Table \ref{table:fpr}. The results show that our proposed FA-Harris method has lower \texttt{FPR} than other methods. eFAST* has low \texttt{FPR} for complex scenes and Arc* has high \texttt{FPR}, which is consistent with \cite{alzugaray2018asynchronous}.  

\subsection{Computational Performance}

\begin{table}[t]
	\centering
	\captionsetup{font={small}}
	\caption{The \iffalse detailed \fi accuracy (\%) of different event-corners detection methods on different scenes. The best results are made in bold. The 'Overall' result here is the weighted average based on the number of events in each scene. }
	\label{table:correct}
	\begin{tabular}{l|ccccc}
	    \toprule[1pt]
		\diagbox{\textbf{Alg.}}{\textbf{Scene}} & \textbf{shapes} & \textbf{dynamic} & \textbf{poster} & \textbf{boxes} & \textbf{Overall} \\
		\hline
		Arc*\cite{alzugaray2018asynchronous}	   & 55.42 & 53.50 & 48.81 & 48.94 & 50.10 \\
		eFAST*\cite{mueggler2017fast}     & 56.22 & 54.86  & 48.30  & 48.60  & 50.06 \\
 		\hline
		eHarris*\cite{vasco2016fast}   & 56.97 & 54.50  & \textbf{49.04}  & 49.26  & 50.56 \\
		G-eHarris*  & 57.04 & 54.50  & 49.02  & 49.35  & 50.57 \\
		\textbf{FA-Harris}  & \textbf{57.66} & \textbf{55.86} & 48.91 & \textbf{49.66} & \textbf{50.88} \\
        \bottomrule[1pt]
	\end{tabular}
\end{table}

\begin{table}[h]
	\centering
	\captionsetup{font={small}}
	\caption{Computational performance of different event-corners detection methods. The best results among Harris-based methods are made in bold.}
	\label{table:comperformance}
	\begin{tabular}{c|cc}
		\toprule[1pt]
		\textbf{Algorithm} & \tabincell{c}{{\textbf{Time per event}}\\{\textbf{[$\mu$s/event]}}} & \tabincell{c}{{\textbf{Max. event rate}} \\{ \textbf{[Meps]}}}\\
		\hline
		Arc*\cite{alzugaray2018asynchronous}	  & 0.14 & 7.23 \\
		eFAST*\cite{mueggler2017fast}    & 0.40 & 2.51 \\
		\hline
		eHarris*\cite{vasco2016fast}  & 5.02 & 0.21 \\
		G-eHarris* & 2.69 & 0.37 \\		
		\textbf{FA-Harris}	  & \textbf{0.66} & \textbf{1.54} \\
		\bottomrule[1pt]
	\end{tabular}
\end{table}

The results of the computational performance for different corner detection methods are shown in Table \ref{table:comperformance}. It reports the average processing time in $\mu$s for a single event and the average rate to process maximum events in Millions of events (Meps) per second for each algorithm, similarly to \cite{alzugaray2018asynchronous}. According to the results, the proposed G-eHarris* method achieves about 2$\times$ speed-up compared to the original eHarris* method. And the proposed FA-Harris corner detector runs about 8$\times$ faster than eHarris*. Unfortunately, FA-Harris is 5$\times$ slower than Arc* method, but we apply the corner candidate refinement method and achieve better accuracy performance as described in \ref{subsec:detector}. 

\section{CONCLUSIONS}
\label{sec:conclusions}
This paper proposes a novel corner detection pipeline (FA-Harris) for event cameras. The proposed detection pipeline includes an event filter,a G-SAE maintaining unit, a corner candidate selecting unit and a corner candidate refining unit. FA-Harris works on a G-SAE and detects the asynchronous event-corners directly without the requirement for event-frames or intensity-frames. The evaluation indicates that FA-Harris runs about 8$\times$ faster than eHarris*. What is more, our FA-Harris also achieves good performance in terms of accuracy. In the future, we will try to use the detected event-corners to perform asynchronous feature tracking with event cameras. We also want to develop an event feature descriptor with scale and rotation invariance. In this way, we can perform fast feature matching and loop closure detection with sparse event-streams.








\section*{ACKNOWLEDGMENT}
\label{sec:acknowledgment}
We are very fortunate to be involved in the project "Unmanaged Cluster System Software Architecture and Operational Support Platform". This project is funded by the National Key R\&D Program of China (2017YFB1001901). This project is also supported by National Natural Science Foundation of China (Grant No. 61903377). Many thanks to Ignacio Alzugaray and Elias Mueggler who shared their code for event-corner detection.

\addcontentsline{toc}{section}{References}
\bibliographystyle{IEEEtran}
\bibliography{IEEEabrv,references}

\end{document}